\documentclass[letterpaper, 10pt, conference]{ieeeconf}  

\IEEEoverridecommandlockouts                              

\usepackage{cite}
\usepackage{amsmath,amssymb,amsfonts}
\usepackage{algorithmic}
\usepackage{graphicx}
\usepackage{textcomp}
\usepackage{xcolor}

\usepackage{acro}			              
\usepackage{bm}			                  
\usepackage{siunitx}		              
\usepackage{hyperref}   	              
\usepackage{booktabs}   	              
\usepackage{tikz, pgfplots, tikz-3dplot}  
\usepackage{xfrac}						  

\pgfplotsset{compat=newest}

\usetikzlibrary{positioning}  
\usetikzlibrary{arrows.meta}  
\usetikzlibrary{calc}         
\usetikzlibrary{decorations.markings}

\newlength\figH
\newlength\figW

\definecolor{TUMBlue}{HTML}{0065BD}            
\definecolor{TUMBlack}{HTML}{000000}           
\definecolor{TUMWhite}{HTML}{ffffff}           
\definecolor{TUMBlueLight}{HTML}{005293}       
\definecolor{TUMBlueDark}{HTML}{003359}        
\definecolor{TUMGrayLight}{HTML}{CCCCCC}       
\definecolor{TUMGray}{HTML}{808080}            
\definecolor{TUMGrayDark}{HTML}{333333}        
\definecolor{TUMAccGray}{HTML}{DAD7CB}         
\definecolor{TUMAccGreen}{HTML}{A2AD00}        
\definecolor{TUMAccOrange}{HTML}{E37222}       
\definecolor{TUMAccBlueDark}{HTML}{64A0C8}     
\definecolor{TUMAccBlueLight}{HTML}{98C6EA}    

\hypersetup{
	pdffitwindow    = false,
	pdfstartview    = FitV,
	pdfnewwindow    = false,
	colorlinks      = true,
	linkcolor       = TUMBlack,
	citecolor       = TUMBlack,
	pdfborder       = {0 0 0},
	urlcolor        = TUMBlack,
}

\DeclareAcronym{3D}{
	short=3D,
	long=three-dimensional
}

\DeclareAcronym{2D}{
	short=2D,
	long=two-dimensional
}

\DeclareAcronym{MPCB}{
	short=MPCB,
	long=Mount Panorama Circuit in Bathurst
}

\DeclareAcronym{LVMS}{
	short=LVMS,
	long=Las Vegas Motor Speedway
}

\DeclareAcronym{RRT}{
	short=RRT,
	long=rapidly exploring random tree
}

\author{Levent Ögretmen$^{1}$, Matthias Rowold$^{1}$, Alexander Langmann$^{2}$, and Boris Lohmann$^{1}$
\thanks{$^{1}$Levent Ögretmen, Matthias Rowold, and Boris Lohmann are with the Chair of Automatic Control, TUM School of Engineering and Design, Technical University of Munich, Germany\newline
	{\tt \{\href{mailto:levent.oegretmen@tum.de}{levent.oegretmen}, \href{mailto:matthias.rowold@tum.de}{matthias.rowold}, \href{mailto:lohmann@tum.de}{lohmann}\}@tum.de}}%
\thanks{$^{2}$Alexander Langmann is with the Professorship of Autonomous Vehicle Systems, TUM School of Engineering and Design, Technical University of Munich, Germany {\tt \href{mailto:alexander.langmann@tum.de}{alexander.langmann@tum.de}}}%
}

\title{\LARGE \bf
	Sampling-Based Motion Planning with Online Racing Line Generation\\
	for Autonomous Driving on Three-Dimensional Race Tracks
}

\newcommand\copyrighttext{%
	\footnotesize \textcopyright 2024 IEEE. Personal use of this material is permitted. Permission from IEEE must be obtained for all other uses, in any current or future media, including reprinting/republishing this material for advertising or promotional purposes, creating new collective works, for resale or redistribution to servers or lists, or reuse of any copyrighted component of this work in other works.
}

\newcommand\copyrightnotice{%
	\begin{tikzpicture}[remember picture,overlay]
		\node[anchor=south,yshift=5pt] at (current page.south) {\fbox{\parbox{\dimexpr\textwidth-\fboxsep-\fboxrule\relax}{\copyrighttext}}};
	\end{tikzpicture}%
}

\begin{document}
	
	\maketitle
	\copyrightnotice
	\thispagestyle{empty}
	\pagestyle{empty}
	
	\begin{abstract}
Existing approaches to trajectory planning for autonomous racing employ sampling-based methods, generating numerous jerk-optimal trajectories and selecting the most favorable feasible trajectory based on a cost function penalizing deviations from an offline-calculated racing line. While successful on oval tracks, these methods face limitations on complex circuits due to the simplistic geometry of jerk-optimal edges failing to capture the complexity of the racing line. Additionally, they only consider two-dimensional tracks, potentially neglecting or surpassing the actual dynamic potential. In this paper, we present a sampling-based local trajectory planning approach for autonomous racing that can maintain the lap time of the racing line even on complex race tracks and consider the race track's three-dimensional effects. In simulative experiments, we demonstrate that our approach achieves lower lap times and improved utilization of dynamic limits compared to existing approaches. We also investigate the impact of online racing line generation, in which the time-optimal solution is planned from the current vehicle state for a limited spatial horizon, in contrast to a closed racing line calculated offline. We show that combining the sampling-based planner with the online racing line generation can significantly reduce lap times in multi-vehicle scenarios.
\end{abstract}
%
	\section{Introduction}
Planning for autonomous racing can be broadly divided into two components: global planning, in which a racing line is planned in the absence of opposing vehicles, and local planning, in which the local environment, i.e., opposing vehicles, is considered. The task of local planning is to adhere to the racing line and, when other vehicles are in the vicinity, make deviations as needed to overtake and prevent collisions.

Sampling-based approaches are commonly employed for local planning due to their simplicity and ability to solve the non-convexity of racing scenarios. In these approaches, multiple trajectory candidates are generated by sampling nodes in the spatio-temporal space and a subsequent generation of edges. The individual trajectories are checked for feasibility, and the optimal one is selected based on a predefined cost function. While sampling-based approaches have been successfully applied to oval race tracks \cite{Raji2022, Ogretmen2022, Ogretmen2024}, following the racing line on a complex circuit in the presence of opposing vehicles presents greater challenges. Furthermore, the existing approaches are limited to \ac{2D} race tracks. However, including \ac{3D} aspects is essential to navigate effectively through elevation changes, banked curves, and various track features. Ignoring these \ac{3D} elements can lead to suboptimal racing performance and hinder the ability to exploit advantageous trajectories on the race track. The racing line from global planning entering the cost function is calculated offline in previous approaches due to the required computation times. However, deviations from the racing line, which regularly occur, for example, due to overtaking maneuvers, result in the racing line losing its optimality, leading to suboptimal overall performance. Emerging approaches in which the racing line is regenerated online hold the potential to immediately respond to deviations from the racing line.

\begin{figure}[t]
	\centering
	\includegraphics[width=1.0\columnwidth]{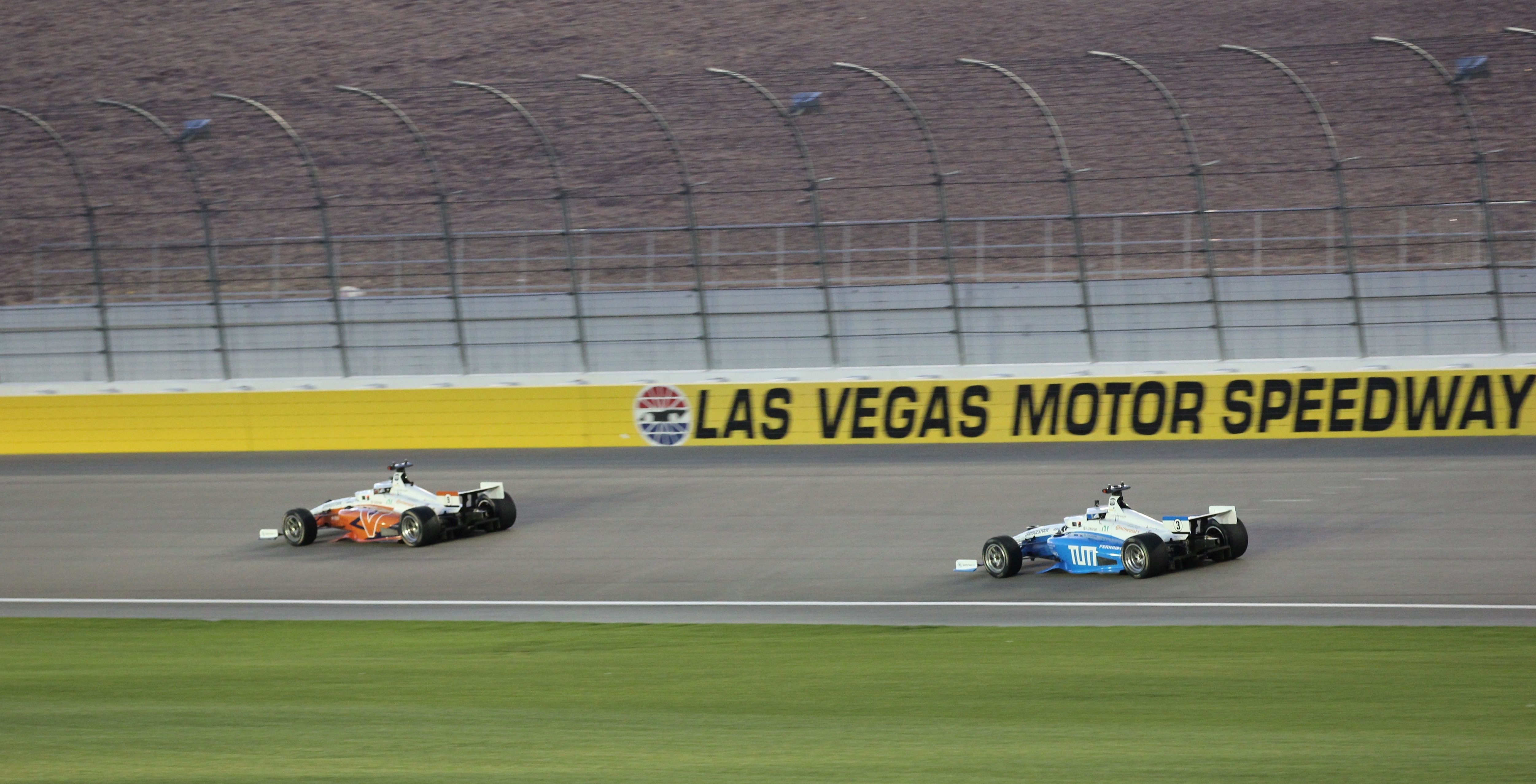}
	\caption{The race cars of the TUM Autonomous Motorsport (right) and Cavalier (left) teams in a banked turn on the \acl{LVMS}.}
	\label{fig:racecars}
\end{figure}

In this paper, we propose the first sampling-based local planning approach that accounts for \ac{3D} effects as they~occur on many race tracks, as shown in Fig.~\ref{fig:racecars}. We further present an adaptation of the trajectory generation of existing approaches, which is necessary for racing line following on complex circuits, enabling lower lap times. Finally, we~evaluate the impact of online racing line generation on the overall performance compared to an offline computed racing line.

\subsection{Related Work}
\label{subsec:related_work}
Numerous resources, like \cite{Claussmann2020}, provide insights into motion planning for traffic situations. A comprehensive examination of global and local planning techniques for racing scenarios can be located in \cite{Betz2022}.

Global planning, which involves the generation of a racing line serving as a reference for local planning, can be performed either offline or online. In the offline approach, a closed racing line is generated, where the initial state corresponds to the final state, representing the optimal trajectory for a complete flying lap. Commonly used optimality criteria for generating the racing line include time \cite{Christ2021} and curvature optimality \cite{Heilmeier2020}. In the online approach, the racing line is regularly re-planned, considering the current vehicle state. Since the computing time must be reduced for the online application, the racing line is not generated for the entire race track but only for a limited horizon, as in \cite{Gundlach2019}. While the aforementioned approach assumes a \ac{2D} race track, Rowold~et~al.~\cite{Rowold2023} perform an online racing line generation considering the effects of a \ac{3D} race track. The \ac{3D} effects are included by constraining the combined lateral and longitudinal acceleration by diamond-shaped gg-diagrams that depend on both velocity and vertical acceleration.

The racing line serves as a reference for local planning, which additionally considers the local environment. Hence, in the event of a slower opposing vehicle occupying the intended racing line, the ego vehicle must deviate from it. The non-convex nature of such an overtaking scenario presents a significant challenge for local planning. While variational methods that solve an optimal control problem by numerical optimization usually find a local optimum depending on the initialization, approaches exist that select the overtaking direction prior to optimization and perform optimization within a constrained sector \cite{Jank2023, Liniger2015}. Other approaches, such as \cite{Ogretmen2024, Rowold2022}, are based on spatio-temporal graphs built along the race track. By performing a graph search, the global discrete-optimal solution can be found. However, due to the exponential increase of the computation time with each dimension, only a coarse discretization of the graph is possible, which limits the application to simple race tracks like oval courses.

Further, sampling-based approaches are often used for local planning. While methods that randomly sample in free space, like \acp{RRT} \cite{LaValle1998}, are generally not used for racing, several approaches exist that generate multiple trajectories by sampling in spatio-temporal space. In \cite{Liniger2020}, a set of trajectories for two vehicles is generated by sampling, and a non-cooperative, nonzero-sum game is solved. Werling~et~al. \cite{Werling2010} generate multiple trajectories for traffic scenarios using jerk-optimal curves in curvilinear coordinates that connect the vehicle state to the sampled nodes. The individual trajectories are checked for feasibility, and the optimal valid one is selected according to a cost function. Raji~et~al. \cite{Raji2022} use these jerk-optimal curves for racing on oval race tracks. In \cite{Ogretmen2022}, the jerk-optimal edges are used to generate the initial edges connecting the vehicle state to a coarse spatio-temporal graph computed offline. The mentioned approaches based on the generation of jerk-optimal curves in curvilinear coordinates are characterized by low computational cost, flexibility in the cost function design, and the ability to resolve non-convexity. However, the jerk-optimal motion primitives are simple in structure and cannot represent complex geometries. This makes it difficult to track the racing line for complex race tracks, so the approaches have been limited to oval race tracks or road traffic. Moreover, none of the mentioned approaches can account for the \ac{3D} effects of a race track.

\subsection{Contribution}
This paper proposes a novel sampling-based local trajectory planning approach for complex \ac{3D} race tracks. We further investigate the effect of online racing line generation on the overall performance. The main contributions can be summarized as follows:
\begin{itemize}
	\item We show that the jerk-optimal trajectory generation used in existing sampling-based planning approaches is sufficient for oval race tracks but unsuitable for complex ones. We adapt the trajectory generation such that the racing line can also be followed on more complex race tracks.
	\item We propose a method for incorporating the \ac{3D} effects of the race track into local planning and show its advantages. Our approach constitutes a generalization of the \ac{2D} sampling-based approaches mentioned in Section~\ref{subsec:related_work}.
	\item In contrast to existing approaches, we do not rely on a fixed offline computed racing line but on an online racing line generation. We investigate the effect of online generation on the overall performance compared to using an offline computed racing line.
\end{itemize}
	\section{Methodology}
\label{sec:methodology}
\def\BoxWidth{4.0cm}
\def\BoxDistance{0.8cm}
\def\HArrowLength{1.3cm}

\tikzset{
	BOX/.style={rectangle, draw=TUMBlueLight, thick, rounded corners, fill=TUMBlueLight!20, very thick, minimum width=\BoxWidth, text width=\BoxWidth, text centered},
	V_ARROW/.style={-Triangle, thick},
	H_ARROW/.style={-Triangle, thick, dashed}
}

\begin{figure}[t]
	\small
	\centering
	\begin{tikzpicture}[
		node distance=\BoxDistance
		]
		
		\node[BOX] (trajectory_generation) {Sampling (Section \ref{sec:sampling}) \\ and trajectory generation \\ (Section \ref{sec:curve_generation})};
		
		\node[BOX] (feasibility_checks) [below=of trajectory_generation] {Feasibility checks \\ (Section \ref{sec:feasibility_checks})};
		
		\node[BOX] (trajectory_selection) [below=of feasibility_checks] {Trajectory selection \\ (Section \ref{sec:trajectory_selection})};
		
		\draw[V_ARROW] (trajectory_generation.south) to node[right] {Set of trajectories} (feasibility_checks.north);
		
		\draw[V_ARROW] (feasibility_checks.south)  to node[right] {Set of valid trajectories} (trajectory_selection.north);
		
		\draw[V_ARROW] (trajectory_selection.south) -- node [right] (trajectory) {Optimal trajectory} ($(trajectory_selection.south)+(0,-\BoxDistance)$);
		
		\draw[H_ARROW] ($(trajectory_generation.east)+(\HArrowLength,0)$) -- node [right, pos=0.0, text width=2.1cm, text centered] (state) {Vehicle state \\ and racing line} (trajectory_generation.east);
		
		\draw[H_ARROW] ($(feasibility_checks.east)+(\HArrowLength,0)$) -- node [right, pos=0.0, text width=2.3cm, text centered] (state) {Physical limits \\ and track bounds} (feasibility_checks.east);
		
		\draw[H_ARROW] ($(trajectory_selection.east)+(\HArrowLength,0)$) -- node [right, pos=0.0, text width=2.1cm, text centered] (state) {Prediction \\ and racing line} (trajectory_selection.east);
		
	\end{tikzpicture}
	\caption{Structure of our local planning approach with all needed inputs.}
	\label{fig:structure}
\end{figure}
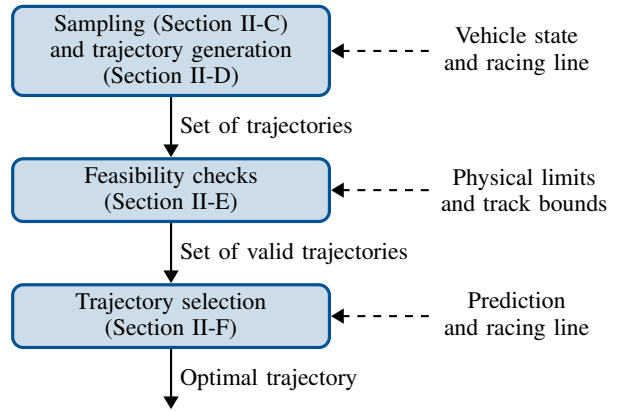
Sections~\ref{sec:track} and \ref{sec:racing_line} introduce the representation of the \ac{3D} race track, the racing line, and all necessary notations. The structure of our local trajectory planning approach is shown in Fig.~\ref{fig:structure} and also serves as the structure for the remainder of this section. The sampling in spatio-temporal space and the connection of the sampled nodes with the vehicle state are described in Sections~\ref{sec:sampling} and \ref{sec:curve_generation}. The feasibility checks of the generated edges follow in Section~\ref{sec:feasibility_checks}, and the selection of the optimal trajectory with the associated definition of the cost function in Section~\ref{sec:trajectory_selection}.

\subsection{3D Track Representation}
\label{sec:track}
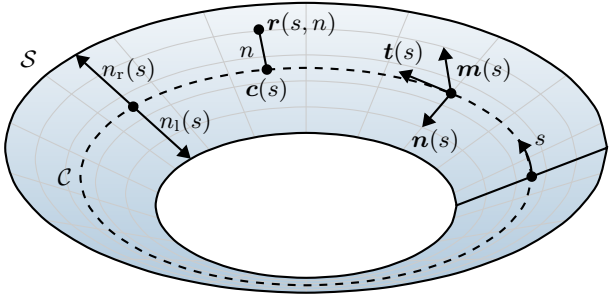
\begin{figure}[t]
	\small
	\centering
	\begin{tikzpicture}[tdplot_main_coords, declare function={a1=4;b1=3;a2=2.0;b2=1.5;h=1.0;}]
		
		\shade[top color=TUMBlueLight!10, bottom color=TUMBlueLight!60, opacity=0.5]
		plot[variable=\x, domain=0:360, smooth] ({a2*cos(\x)}, {b2*sin(\x)}, 0)
		-- plot[variable=\x, domain=360:0, smooth] ({a1*cos(\x)}, {b1*sin(\x)}, h);
		
		\foreach \heightscale in {0.2, 0.4, 0.6, 0.8}{
			\draw[thin, black!20] 
			plot[variable=\x,domain=0:360,smooth]
			({(\heightscale*(a1-a2)+a2)*cos(\x)},{(\heightscale*(b1-b2)+b2)*sin(\x)},\heightscale*h);
		}
		
		\foreach \angle in {10,30,...,360}{
			\draw[thin, black!20] 
			({a2*cos(\angle)},{b2*sin(\angle)},0) -- ({a1*cos(\angle)},{b1*sin(\angle)},h);
		}
		
		\draw[thick] 
		plot[variable=\x,domain=0:360,smooth] ({a2*cos(\x)},{b2*sin(\x)},0)
		-- plot[variable=\x,domain=360:0,smooth] ({a1*cos(\x)},{b1*sin(\x)},h);
		
		\draw[dashed, thick] 
		plot[variable=\x,domain=0:360,smooth]
		({(a1+a2)/2*cos(\x)},{(b1+b2)/2*sin(\x)},h/2);
		
		\draw[{Circle[length=4pt]}-Triangle, thick, shorten <=-2pt]
		({(a1+a2)/2*cos(50)},{(b1+b2)/2*sin(50)},h/2) 
		-- node [above, pos=0.9] {$\bm{t}(s)$} 
		($({(a1+a2)/2*cos(50)},{(b1+b2)/2*sin(50)},h/2)+(-0.7, 0.45, 0)$);
		
		\draw[-Triangle, thick]
		({(a1+a2)/2*cos(50)},{(b1+b2)/2*sin(50)},h/2) 
		-- node [below, pos=0.8, xshift=0.1cm] {$\bm{n}(s)$} 
		({(0.2*(a1-a2)+a2)*cos(50)},{(0.2*(b1-b2)+b2)*sin(50)},0.2*h);
		
		\draw[-Triangle, thick]
		({(a1+a2)/2*cos(50)},{(b1+b2)/2*sin(50)},h/2) 
		-- node [right] {$\bm{m}(s)$} 
		($({(a1+a2)/2*cos(50)},{(b1+b2)/2*sin(50)},h/2)+(-0.1, 0.6, 0.3)$);
		
		\draw[{Circle[length=4pt]}-{Circle[length=4pt]}, thick, shorten <=-2pt]
		({(a1+a2)/2*cos(100)},{(b1+b2)/2*sin(100)},h/2) 
		-- node [below, pos=0.0] {$\bm{c}(s)$} 
		node [left, pos=0.4] {$n$} 
		node [right, pos=1.0] {$\bm{r}(s,n)$}
		({(0.85*(a1-a2)+a2)*cos(100)},{(0.85*(b1-b2)+b2)*sin(100)},0.85*h);
		
		\draw[{Circle[length=4pt]}-Triangle, thick, shorten <=-2pt] plot[variable=\x,domain=0:20,smooth] ({(a1+a2)/2*cos(\x)},{(b1+b2)/2*sin(\x)},h/2) 
		node[right, xshift=0.1cm] {$s$};
		
		\draw[{Circle[length=4pt]}-Triangle, thick, shorten <=-2pt]
		({(a1+a2)/2*cos(140)},{(b1+b2)/2*sin(140)},h/2) 
		-- node [right, pos=0.7] {$n_\mathrm{r}(s)$} ({a1*cos(140)},{b1*sin(140)},h);
		
		\draw[-Triangle, thick] 
		({(a1+a2)/2*cos(140)},{(b1+b2)/2*sin(140)},h/2) 
		-- node [right, pos=0.3] {$n_\mathrm{l}(s)$} ({a2*cos(140)},{b2*sin(140)},0);
		
		\node[above left] at ({a1*cos(150)},{b1*sin(150)},h) {$\mathcal{S}$};
		
		\node[left] at ({(a1+a2)/2*cos(180)},{(b1+b2)/2*sin(180)},h/2) {$\mathcal{C}$};
		
	\end{tikzpicture}
	\caption{Representation of a closed, counterclockwise orientated \ac{3D} race track with negative banking ($\varphi<0$) and without slope ($\mu=0$).}
	\label{fig:3dtrack}
\end{figure}

For the 3D race track, we use the representation introduced in \cite{Perantoni2015}, which is based on a spine and a lateral offset to it. The spine, referred to in the following as the reference line, is a three-dimensional curve
\begin{equation}
	\mathcal{C} = \left\{ \bm{c}(s) \in \mathbb{R}^3 \mid s \in [0, s_{\mathrm{f}}] \right\}
	\label{eq:spine}
\end{equation}
parameterized with respect to its arc length $s$. Since we are specifically considering closed race tracks, it holds that $\bm{c}(0)=\bm{c}(s_{\mathrm{f}})$. We now introduce a moving coordinate system associated with the reference line $\mathcal{C}$, as depicted in Fig.~\ref{fig:3dtrack}. This coordinate system, which we refer to as the road frame, is spanned by the three unit vectors $\bm{t}(s)$, $\bm{n}(s)$, and $\bm{m}(s)$. The unit vector $\bm{t}(s)=\frac{\mathrm{d} \bm{c}(s)}{\mathrm{d} s}$ is tangential to the reference line $\mathcal{C}$. $\bm{n}(s)$ is orthogonal to $\bm{t}(s)$ and lies in the plane of the race track. The road frame is completed with $\bm{m}(s)=\bm{t}(s)\times\bm{n}(s)$, which is thus orthogonal. The orientation of the coordinate system is expressed by intrinsic zyx Euler angles, where $\theta(s)$, $\mu(s)$, and $\varphi(s)$ refer to the rotations around the corresponding axes. The angular velocity of the road frame with respect to the arc length $s$ is denoted as $\bm{\Omega}(s) = \begin{bmatrix} \Omega_{\mathrm{x}}(s) & \Omega_{\mathrm{y}}(s) & \Omega_{\mathrm{z}}(s) \end{bmatrix}^\intercal$ and can be derived from the derivatives of the Euler angles \cite{Rowold2023}. Given the width between the reference line and the left and right track boundaries $n_{\mathrm{l}}(s)$ and $n_{\mathrm{r}}(s)$, the race track surface $\mathcal{S}$ can now be defined as:
\begin{equation}
	\begin{aligned}
		\mathcal{S} = \left\{ \bm{r}(s, n) = \bm{c}(s) \right. &+ \bm{n}(s) \cdot n \in \mathbb{R}^3 \mid \\
		& \left. s \in [0, s_{\mathrm{f}}], n \in [n_{\mathrm{r}}(s), n_{\mathrm{l}}(s)] \right\}.
	\end{aligned}
\end{equation}
While $s$ represents the progress along the reference line of length $s_{\mathrm{f}}$, $n$ corresponds to the lateral offset on the plane of the race track relative to the reference line.

\tdplotsetmaincoords{50}{0} 

\subsection{Racing Line}
\label{sec:racing_line}
The racing line $\mathcal{R}$ with a fixed spatial horizon $H_{\mathrm{rl}}$ and a respective time horizon $T_{\mathrm{rl}}$ is given as a curve parameterized with respect to time $t$ in curvilinear coordinates:
\begin{equation}
	\mathcal{R} = \left\{ \left( s_{\mathrm{rl}}(t), n_{\mathrm{rl}}(t) \right) \mid t \in \left[ 0, T_{\mathrm{rl}} \right] \right\}.
\end{equation}
We assume here that the racing line originates at the current longitudinal vehicle position $s_{\mathrm{0}}$, i.e., $s_{\mathrm{rl}}(0) = s_{\mathrm{0}}$ holds. The sampling in Section~\ref{sec:sampling} and the trajectory generation in Section~\ref{sec:curve_generation} are performed depending on $s_{\mathrm{rl}}(t)$, $n_{\mathrm{rl}}(t)$, and their temporal derivatives $\dot{s}_{\mathrm{rl}}(t)$, $\ddot{s}_{\mathrm{rl}}(t)$, $\dot{n}_{\mathrm{rl}}(t)$, and $\ddot{n}_{\mathrm{rl}}(t)$ to generate trajectories with a tendency towards the racing line. However, it should also be possible to follow the path of the racing line with a longitudinal profile $s(t)$ deviating from $s_{\mathrm{rl}}(t)$, e.g., in case of a braking maneuver. In this case, the lateral profile $n_{\mathrm{rl}}(t)$ is traversed with a different temporal profile, which requires adjustment of its time derivatives.

For a given longitudinal curve $s_i(t)$, the lateral profile $\tilde{n}_{\mathrm{rl},i}(t)$ required to replicate the path of the racing line can be stated as the function composition $\tilde{n}_{\mathrm{rl},i}(t) = n_{\mathrm{rl}} \left( s_{\mathrm{rl}}^{-1} \left( s_i\left( t \right) \right) \right)$. Applying the chain and inverse function rules to the emergent derivatives of the inverse function $s_{\mathrm{rl}}^{-1} \left( s_i \left( t \right) \right)$, we obtain the time derivatives, omitting the arguments for brevity:
\begin{subequations}
	\begin{equation}
		\dot{\tilde{n}}_{\mathrm{rl},i} = \frac{ \dot{n}_{\mathrm{rl}} }{ \dot{s}_{\mathrm{rl}} } \cdot \dot{s}_i,
	\end{equation}
	\begin{equation}
		\ddot{\tilde{n}}_{\mathrm{rl},i} = \left( \frac{ \ddot{n}_{\mathrm{rl}} }{ \dot{s}_{\mathrm{rl}}^2 }
		- \frac{ \dot{n}_{\mathrm{rl}} \ddot{s}_{\mathrm{rl}}}{ \dot{s}_{\mathrm{rl}}^3 } \right) \cdot \dot{s}_i^2
		+ \frac{ \dot{n}_{\mathrm{rl}} }{ \dot{s}_{\mathrm{rl}} } \cdot \ddot{s}_i.
	\end{equation}
\end{subequations}

\subsection{Sampling in Curvilinear Coordinates}
\label{sec:sampling}
Given the vehicle state $S_{\mathrm{0}} = \left[s_{\mathrm{0}}, \dot{s}_{\mathrm{0}}, \ddot{s}_{\mathrm{0}}, n_{\mathrm{0}}, \dot{n}_{\mathrm{0}}, \ddot{n}_{\mathrm{0}}\right]$, we sample end nodes $S_{\mathrm{e},ij} = \left[\dot{s}_{\mathrm{e},i}, \ddot{s}_{\mathrm{e},i}, n_{\mathrm{e},j}, \dot{n}_{\mathrm{e},j}, \ddot{n}_{\mathrm{e},j}\right]$ and then connect $S_{\mathrm{0}}$ to the sampled nodes $S_{\mathrm{e},ij}$ with a fixed horizon $T$, resulting in a set of trajectories. As the computational cost increases exponentially with each dimension, not all quantities are sampled in $S_{\mathrm{e},ij}$, but only the longitudinal velocity $\dot{s}_{\mathrm{e}, i}$ and the lateral position $n_{\mathrm{e},j}$. The higher-order derivatives, on the other hand, are selected appropriately without further sampling, as explained below. Moreover, $S_{\mathrm{e},ij}$ does not include a longitudinal end position, as in the racing scenario the selected end velocities $\dot{s}_{\mathrm{e},i}$ are more decisive and the end positions adjust accordingly.

Since an essential task of local planning is to follow the racing line $\mathcal{R}$, the sampled nodes $S_{\mathrm{e},ij}$ must be chosen so that the resulting trajectories can represent the racing line. In the longitudinal direction, $N_{\mathrm{\dot{s}}}$ equidistantly distributed samples are generated in the interval $\dot{s}_{\mathrm{e},i} \in \left[ 0, \dot{s}_{\mathrm{rl}}\left(T\right) \cdot K_{\mathrm{\dot{s}}} \right]$, additionally always including the terminal racing line velocity $\dot{s}_{\mathrm{rl}}\left(T\right)$. The enlargement of the range by $K_{\mathrm{\dot{s}}}>1$ is required to allow for possible necessary deviation from the racing line. The acceleration $\ddot{s}_{\mathrm{e},i}$ associated with the sampled $\dot{s}_{\mathrm{e},i}$ is chosen to be $\ddot{s}_{\mathrm{rl}}(T)$ to allow tracking of the racing line. However, to obtain reasonable trajectories also for larger deviations from the racing line, we interpolate linearly between $0$ and $\ddot{s}_{\mathrm{rl}}(T)$ depending on the proximity of $\dot{s}_{\mathrm{0}}$ to $\dot{s}_{\mathrm{rl}}(0)$ and $\dot{s}_{\mathrm{e},i}$ to $\dot{s}_{\mathrm{rl}}(T)$.

Given a longitudinal curve $s_i(t)$, we generate several lateral curves that complement it. For this, we sample $N_{\mathrm{n}}$ equidistantly distributed samples within the track boundaries $n_{\mathrm{e},j} \in \left[ n_{\mathrm{r}}(s_{\mathrm{e},i}) + \sfrac{d_{\mathrm{w}}}{2}, \, n_{\mathrm{l}}(s_{\mathrm{e},i}) - \sfrac{d_{\mathrm{w}}}{2} \right]$, considering the vehicle width $d_{\mathrm{w}}$ and again including the terminal racing line position $\tilde{n}_{\mathrm{rl},i}(T)$. We select the corresponding time derivatives $\dot{n}_{\mathrm{e},j}$ and $\ddot{n}_{\mathrm{e},j}$ to ensure the alignment at the track boundaries and the racing line and interpolate linearly in between. Note that the lateral samples and curves must be recalculated for each $s_i(t)$, as $\tilde{n}_{\mathrm{rl},i}(t)$ changes accordingly.

\subsection{Trajectory Generation}
\label{sec:curve_generation}
For trajectory generation, the initial state $S_{\mathrm{0}}$ is connected to the sampled end nodes $S_{\mathrm{e},ij}$, and the resulting $s_i(t)$ and $n_j(t)$ pairs are transformed into Cartesian coordinates for the subsequent feasibility checks. First, we examine the creation of jerk-optimal trajectories as described in existing literature. Following that, we introduce our novel approach to trajectory generation, offering enhanced racing line tracking capabilities on complex race tracks.

As shown in \cite{Takahashi1989}, for a point mass, quintic polynomials connecting a start state $[p_{\mathrm{0}}, \dot{p}_{\mathrm{0}}, \ddot{p}_{\mathrm{0}}]$ to an end state $[p_{\mathrm{e}}, \dot{p}_{\mathrm{e}}, \ddot{p}_{\mathrm{e}}]$ with the time horizon $T$ minimize the cost functional
\begin{equation}
	J(p(t)) = \frac{1}{2} \int_{0}^{T} \dddot{p}^2(t) \, \mathrm{d} t
	\label{eq:jerkcost}
\end{equation}
and are therefore jerk-optimal. If $p_{\mathrm{e}}$ is arbitrary, causing the end state to lie on a manifold, quartic polynomials minimize the cost functional \eqref{eq:jerkcost} as derived in \cite{Werling2012}. Applied to our case, quintic polynomials can be used for curve generation in the lateral direction and quartic polynomials in the longitudinal direction. The resulting jerk-optimal curves in curvilinear coordinates reduce abrupt acceleration changes, which has a beneficial effect on the subsequent tracking control as well as vehicle stability and are used in the context of the racing scenario in \cite{Raji2022} and \cite{Ogretmen2022}. On the other hand, these trajectories exhibit only a simple structure, which generally cannot represent the racing line geometry for a complex race track. This leads to a degraded racing line tracking performance so that the jerk-optimal trajectories can only be applied reasonably for race tracks with simple racing lines that can be approximated well by the polynomials, such as oval race tracks. The racing line tracking in the lateral direction can be improved by selecting a reference line closer to the racing line to be driven. However, such a measure is unsuitable for online racing line generation due to the varying racing line. Moreover, the problem of tracking in the longitudinal direction remains, requiring an adaptation of the trajectory generation.

The main idea is to generate the jerk-optimal curves relative to the racing line and then add the racing line to them. This allows the racing line to be tracked exactly and simultaneously enables smooth jerk-optimal deviations from the racing line, e.g., during overtaking or braking maneuvers. For realization, we subtract the racing line start conditions from the start state
\begin{equation}
	\begin{aligned}
		\tilde{S}_{\mathrm{0}} = \left[s_{\mathrm{0}} \right. & - s_{\mathrm{rl}}(0), \dot{s}_{\mathrm{0}} - \dot{s}_{\mathrm{rl}}(0), \ddot{s}_{\mathrm{0}} - \ddot{s}_{\mathrm{rl}}(0), \\
		&\left. n_{\mathrm{0}} - \tilde{n}_{\mathrm{rl},i}(0), \dot{n}_{\mathrm{0}} - \dot{\tilde{n}}_{\mathrm{rl},i}(0), \ddot{n}_{\mathrm{0}} - \ddot{\tilde{n}}_{\mathrm{rl},i}(0) \right]
	\end{aligned}
\end{equation}
and the end conditions from the sampled end state 
\begin{equation}
	\begin{aligned}
		\tilde{S}_{\mathrm{e},ij} = \left[\dot{s}_{\mathrm{e},i} - \dot{s}_{\mathrm{rl}}(T)\right. &, \ddot{s}_{\mathrm{e},i} - \ddot{s}_{\mathrm{rl}}(T), n_{\mathrm{e},j} - \tilde{n}_{\mathrm{rl},i}(T), \\
		&\left. \dot{n}_{\mathrm{e},j} - \dot{\tilde{n}}_{\mathrm{rl},i}(T), \ddot{n}_{\mathrm{e},j} - \ddot{\tilde{n}}_{\mathrm{rl},i}(T) \right]
	\end{aligned}
\end{equation}
and connect the adapted states with the mentioned quartic and quintic polynomials. To comply with the actual start state $S_{\mathrm{0}}$ and the sampled end states $S_{\mathrm{e},ij}$ specified in Section~\ref{sec:sampling}, the racing line profiles are then added to the resulting curves. Fig.~\ref{fig:jerk_vs_relative} shows exemplarily the comparison of jerk-optimal curves and curves generated relative to the racing line, referred to as relative curves in the following. It shows the tendency of the relative edges to the racing line and that, in contrast to the jerk-optimal curves, the basic geometry of the racing line can be recreated.

\setlength{\figH}{0.57\columnwidth}
\setlength{\figW}{1.0\columnwidth}
\begin{figure}[t]
	\centering
	\small
	\input{figures/curve_comparison.tex}
	\caption{Comparison of jerk-optimal and relative trajectory generation with $\dot{s}_{\mathrm{0}} = \SI{53}{\metre\per\second}$, $\ddot{s}_{\mathrm{0}} = \SI{2}{\metre\per\second\squared}$, $N_{\mathrm{\dot{s}}} = \num{6}$, and $T = \SI{4}{\second}$.}
	\label{fig:jerk_vs_relative}
\end{figure}

The relative trajectory generation can track the racing line if the vehicle velocity is close to the racing line velocity, but problems arise if the vehicle velocity deviates strongly from it. Specifically, reaching the racing line velocity is only possible with restrictions, as the course of the trajectories depends on the position on the race track. For example, at a section of the race track with a strong deceleration of the racing line, there may be no longitudinal curve with a monotonically increasing velocity profile, which prevents a proper approach to the racing line velocity at this location. To address this issue, we switch to the jerk-optimal generation for the longitudinal curves if $| \dot{s}_{\mathrm{0}} - \dot{s}_{\mathrm{rl}}(0) | \, / \,  \dot{s}_{\mathrm{rl}}(0)$ exceeds a threshold $s_{\mathrm{0},\mathrm{thr}}$. In the lateral direction, however, we use both the jerk-optimal and the relative curves. Thus, for every longitudinal profile $s_i(t)$, two lateral curves are generated for each sampled end position $n_{\mathrm{e},j}$. This increases the variety of maneuvers that can be planned, which is particularly advantageous for overtaking maneuvers.

After generating the curves in curvilinear coordinates, these are converted into Cartesian coordinates. Analogous to the derivation in \cite{Werling2010}, we perform the closed transformation $\left[ s, \dot{s}, \ddot{s}, n, \dot{n}, \ddot{n} \right](t) \rightarrow \left[ x, y, z, v, \hat{\chi}, \hat{a}_{\mathrm{x}}, \hat{a}_{\mathrm{y}}, \hat{\kappa} \right](t)$ for this. In contrast to \cite{Werling2010}, in which only \ac{2D} tracks are considered, the variables here refer to the newly introduced vehicle's velocity coordinate system on the \ac{3D} road plane. This velocity frame is obtained by rotating the road frame introduced in Section~\ref{sec:track} around the $\bm{m}(s)$ axis (z-axis) by the angle $\hat{\chi}$ such that the x-axis is parallel to the projection of the absolute vehicle velocity onto the road plane $v$. The origin of the velocity coordinate system is formed by $x$, $y$, and $z$, representing the geometric center of the vehicle on the road plane. $\hat{a}_{\mathrm{x}}$ and $\hat{a}_{\mathrm{y}}$ are the accelerations in the x- and y-axis, while $\hat{\kappa}$ corresponds to the curvature on the road plane.

\subsection{Feasibility Checks}
\label{sec:feasibility_checks}
We perform three checks on each generated trajectory to verify its feasibility. First, the path of the trajectory is examined to ensure that it remains within the track bounds, taking into account the vehicle geometry. To avoid a costly check in Cartesian coordinates, we carry out a simplified check in curvilinear coordinates with the vehicle width $d_{\mathrm{w}}$:
\begin{equation}
	n_{\mathrm{r}}(s(t)) + \frac{d_{\mathrm{w}}}{2} + d_{\mathrm{s}} \leq n(t) \leq n_{\mathrm{l}}(s(t)) - \frac{d_{\mathrm{w}}}{2} - d_{\mathrm{s}}.
	\label{eq:pathcheck}
\end{equation}
The check is based on the assumption that the vehicle drives parallel to the reference line, which can be compensated for by adding a safety margin $d_{\mathrm{s}}$. Secondly, as a kinematic condition, we check whether the minimal turning radius of the vehicle $r_{\mathrm{min}}$ is complied with: $\hat{\kappa}(t) \leq \sfrac{1}{r_{\mathrm{min}}} = \hat{\kappa}_{\mathrm{max}}$.

Lastly, compliance with the dynamic limits of the vehicle, i.e., the combined longitudinal and lateral acceleration, must be ensured. For an efficient check, we use gg-diagrams that constrain the combined accelerations of the vehicle and are calculated offline. These represent the complexity of a sophisticated dynamic vehicle model and are based on the quasi-steady-state assumption that each point along a trajectory corresponds to stationary circular motion without considering transient effects. While the gg-diagrams only depend on velocity $v$ on \ac{2D} tracks, they can be extended for \ac{3D} tracks. For this purpose, we enhance the gg-diagrams by the dependence of the vertical acceleration $\tilde{g}$ as in \cite{Lovato2022}. This acceleration depends on the vehicle's movement along the \ac{3D} track and may, therefore, deviate from the gravitational acceleration $g = \SI{9.81}{\meter\per\second\squared}$. For the generation of the gg-diagrams, we follow the method introduced in \cite{Rowold2023}, in which a computationally advantageous diamond-shaped underapproximation of the actual diagram shape is derived. An underestimation is necessary as the actual vehicle limits should not be exceeded. As in \cite{Veneri2020}, the actual limit shape is obtained based on a double-track model and a Pacejka tire model. The diamond-shaped underapproximation is characterized by its vertices $\tilde{a}_{ \mathrm{x}, \mathrm{min} }(v, \tilde{g})$ and $\tilde{a}_{ \mathrm{y}, \mathrm{max} }(v, \tilde{g})$ as well as a shape factor $1 \leq p(v, \tilde{g}) \leq 2$. By varying $p$, the shape varies from a rhombus ($p=1$) to an ellipse ($p=2$). The final fourth parameter is $\tilde{a}_{ \mathrm{x}, \mathrm{max} }(v, \tilde{g})$, limiting the positive longitudinal acceleration, meaning no symmetry is given in the longitudinal direction. Overall, the check is carried out using inequalities \eqref{eq:diamond1} to \eqref{eq:diamond3}, with the time arguments being neglected for reasons of clarity.
\begin{subequations}
	\begin{align}
		\tilde{a}_{\mathrm{x}} & \leq \tilde{a}_{\mathrm{x}, \mathrm{max} }(v, \tilde{g}) \label{eq:diamond1} \\
		\left|\tilde{a}_{\mathrm{y}}\right| & \leq \tilde{a}_{\mathrm{y}, \mathrm{max} }(v, \tilde{g}) \\
		\left|\tilde{a}_{\mathrm{x}}\right| & \leq\left|\tilde{a}_{\mathrm{x}, \mathrm{min} }(v,\tilde{g})\right| \left[1-\left[\frac{\left|\tilde{a}_{\mathrm{y}}\right|}{\tilde{a}_{\mathrm{y}, \mathrm{max} }(v, \tilde{g})}\right]^{p(v, \tilde{g})}\right]^{\frac{1}{p(v, \tilde{g})}} \label{eq:diamond3}
	\end{align}
\end{subequations}
Note that in these inequalities the apparent accelerations $\tilde{a}_{\mathrm{x}}(t)$ and $\tilde{a}_{\mathrm{y}}(t)$, which include the effect of the gravitational acceleration, are constrained instead of the accelerations $\hat{a}_{\mathrm{x}}(t)$ and $\hat{a}_{\mathrm{y}}(t)$ introduced in Section~\ref{sec:curve_generation}. The correlations required for the evaluation of \eqref{eq:diamond1} to \eqref{eq:diamond3}, neglecting the arguments, are
\begin{equation}
	\left[\begin{array}{c}
		\tilde{a}_{\mathrm{x}} \\
		\tilde{a}_{\mathrm{y}} \\
		\tilde{g}
	\end{array}\right]
	=
	\left[\begin{array}{c}
		\hat{a}_{\mathrm{x}}+g\left(\mathrm{c}_{\mu} \mathrm{s}_{\varphi} \mathrm{s}_{\hat{\chi}}-\mathrm{s}_{\mu} \mathrm{c}_{\hat{\chi}}\right) \\
		\hat{a}_{\mathrm{y}}+g\left(\mathrm{s}_{\mu} \mathrm{s}_{\hat{\chi}}+\mathrm{c}_{\mu} \mathrm{s}_{\varphi} \mathrm{c}_{\hat{\chi}}\right) \\
		\dot{w}-\hat{\omega}_{\mathrm{y}} v+g\left(\mathrm{c}_{\mu} \mathrm{c}_{\varphi}\right)
	\end{array}\right].
	\label{eq:hat_tilde}
\end{equation}
Here, $\mathrm{s}_\square$ and $\mathrm{c}_\square$ are abbreviations for $\sin\square$ and $\cos\square$, and the terms negligible according to \cite{Rowold2023} are omitted. In addition, the time derivative of the vertical velocity $\dot{w} = \dot{n} \omega_{\mathrm{x}} + n \dot{\omega}_{\mathrm{x}}$ with $\omega_{\mathrm{x}} = \Omega_{\mathrm{x}} \dot{s}$ and the angular velocity $\hat{\omega}_{\mathrm{y}} = \left( \Omega_{\mathrm{y}} c_{\hat{\chi}} - \Omega_{\mathrm{x}} s_{\hat{\chi}} \right) \dot{s}$ of the velocity frame appear. A full derivation of \eqref{eq:hat_tilde} is given in the appendix of \cite{Rowold2023}.

Further, we apply a soft check concept for all three introduced checks. If all trajectories within a planning cycle are not feasible in one of the three checks, the trajectory with the fewest violations is selected. Consequently, a solution is always available that brings the vehicle into a viable state.

An essential aspect of the introduced checks is the consistency of the limits between the racing line and the local planning. To enable the racing line to be tracked, it must comply with the checks performed by the local planner. Moreover, a margin for the limits must be included in the local planning compared to the racing line, e.g., to provide additional acceleration potential needed for overtaking maneuvers. To realize this additional potential for local planning, we choose a larger safety margin $d_{\mathrm{s},\mathrm{rl}} > d_{\mathrm{s}}$ for the racing line in \eqref{eq:pathcheck} and also scale the extremal values $\tilde{a}_{\mathrm{x}, \mathrm{min} }(v, \tilde{g})$, $\tilde{a}_{\mathrm{x}, \mathrm{max} }(v, \tilde{g})$ and $\tilde{a}_{\mathrm{y}, \mathrm{max} }(v, \tilde{g})$ for the racing line in \eqref{eq:diamond1} to \eqref{eq:diamond3} with $1-\tilde{a}_{\mathrm{mgn}}$, where $\tilde{a}_{\mathrm{mgn}} \in [0, 1)$.

Apart from the three checks, we do not perform a collision check with opposing vehicles. Instead, other vehicles are avoided due to an additional term in the cost function introduced in Section~\ref{sec:trajectory_selection}. This is sufficient for the scenarios evaluated in this paper and has the advantage of a reduced computation time and that an infeasible state due to opposing vehicles within the safety distance is avoided.

\subsection{Trajectory Selection}
\label{sec:trajectory_selection}
\begin{figure}[t]
	\centering
	\small
	\begin{tikzpicture}[declare function={ks=0.015;kn=0.5;}]
		\begin{axis}[
			view={45}{20},
			width=1.1\columnwidth,
			enlargelimits=false,
			axis equal image,
			unit vector ratio=1 1 7,
			grid,
			xmin=-15, xmax=15,
			ymin=-5, ymax=5,
			zmin=0, zmax=1,
			domain=-15:15,
			y domain=-5:5,
			colormap={tumcolormap}{color(0cm)=(TUMBlue!70); color(0.4cm)=(TUMGrayLight); color(1cm)=(TUMAccOrange)}, 
			point meta max=1, 
			point meta min=0,
			xlabel=$s$ in \si{\meter},
			ylabel=$n$ in \si{\meter},
			zlabel=cost,
			xlabel style={rotate=-20, xshift=0.0cm, yshift=0.3cm},
			ylabel style={rotate=25, xshift=0.0cm, yshift=0.2cm},
			ztick = {0.0, 0.5, 1.0},
			]
			
		\newcommand\expr[2]{ exp(-ks * #1^2 - kn * #2^2) }

		
		\addplot3 [
		contour gnuplot={contour dir=y,draw color=black!50,labels=false,levels={0, 0.4, 0.8, 1.2, 1.6, 2, 2.4, 2.8}},
		y filter/.expression={5},
		samples=60,
		] {\expr{x}{y}};
		
		
		\addplot3 [
		contour gnuplot={contour dir=x,draw color=black!50,labels=false,levels={0, 2, 4, 6, 8, 10, 12, 14}},
		x filter/.expression={-15},
		samples=60,
		] {\expr{x}{y}};
		
		
		\addplot3 [
		surf,
		samples=60,
		] {\expr{x}{y}};
		
		
		\end{axis}
	\end{tikzpicture}
	\caption{Elliptical shaped prediction costs with $k_\mathrm{s}=\num{0.015}$ and $k_\mathrm{n}=\num{0.5}$ of an opposing vehicle located at $s_{\mathrm{pr},m}(t)=n_{\mathrm{pr},m}(t)=0$.}
	\label{fig:prediction_cost}
\end{figure}
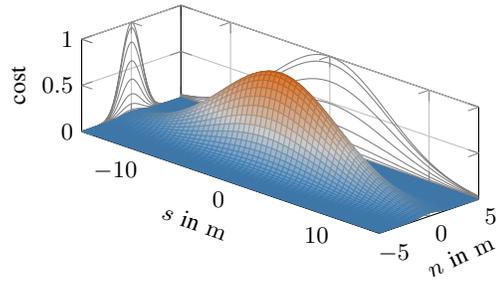
In the last step, costs are assigned to all feasible trajectories, and the cost-minimal trajectory is selected. The design of the cost function must be chosen so that the racing line is followed as closely as possible and, if necessary, other vehicles are avoided. This requires a prediction $\left( s_{\mathrm{pr},m}(t), n_{\mathrm{pr},m}(t) \right), \, t \in \left[ 0, T \right]$ for each considered opposing vehicle indexed with $m$, which is assumed to be given. The scalar total cost $C$ of a specific edge is calculated by integration over time
\begin{equation}
	C = \int_{0}^{T} w_{\mathrm{n}} \cdot \Delta n^2(t) + w_{\mathrm{v}} \cdot \frac{\Delta v^2(t)}{v_{\mathrm{rl}}^2(t)} + w_{\mathrm{pr}} \cdot d_{\mathrm{pr}}(t) \,\, \mathrm{d} t,
	\label{eq:cost:function}
\end{equation}
where the terms constitute a trade-off between racing line following and prediction avoidance, with $\Delta n(t) = n_{\mathrm{rl}}(t) - n(t) $ and $\Delta v(t) = v_{\mathrm{rl}}(t) - v(t)$. The parameters $w_\square$ allow individual weighting of each term.

The first two terms penalize the lateral position deviation and the relative velocity deviation to the racing line, thereby achieving adherence to the racing line in the absence of opposing vehicles. The third term penalizes trajectories close to the prediction of the opposing vehicles, thus enabling overtaking maneuvers while maintaining a sufficient distance. $d_{\mathrm{pr}}(t)$ is computed using the exponential expression
\begin{equation}
	d_{\mathrm{pr}}(t) = \sum_{m=0}^{M-1}e^{- k_{\mathrm{s}} \left[ s_{\mathrm{pr},m}(t) - s(t) \right]^2 - k_{\mathrm{n}} \left[ n_{\mathrm{pr},m}(t) - n(t) \right]^2 },
	\label{eq:prediction_cost}
\end{equation}
where $M$ expresses the current number of opposing vehicles within the sensor range $d_{\mathrm{snr}}$. It spans an elliptical shape around the predicted vehicle center, as depicted in Fig.~\ref{fig:prediction_cost}. The shape is parameterized by $k_{\mathrm{s}}$ and $k_{\mathrm{n}}$, which allows an overapproximation of the actual vehicle geometry and a safety distance. The ellipse has a maximum value of \num{1} in the geometric center of the vehicle and tends exponentially towards \num{0} as the distance increases. Due to the exponential progression, no sharp transition exists, preventing sudden changes in the trajectory planning between two planning cycles. In contrast to the cost design in \cite{Rowold2022}, where also an elliptical shape is used, the ellipses in \eqref{eq:prediction_cost} are not aligned with the heading of the opposing vehicles due to the formulation in curvilinear coordinates. This prevents blockage of the entire track width by the cost ellipse, which may otherwise occur in the entry of sharp turns or due to uncertainties in the heading estimation of opponent vehicles.
	\section{Results and Discussion}
\label{sec:results_discussion}
\begin{figure}%
	\small
	\begin{minipage}[t]{0.355\columnwidth}
		\setlength{\figH}{2.8cm}
\begin{tikzpicture}

\definecolor{darkcyan0101189}{RGB}{0,101,189}
\definecolor{darkgray176}{RGB}{176,176,176}

\begin{axis}[
height=\figH,
scale only axis,
tick pos=both,
width=1.0815578137044577*\figH,
x grid style={darkgray176},
xmajorgrids,
xmin=3.02621289999999, xmax=1006.8124861,
xtick style={color=black},
xtick={0,200,...,1000},
xticklabels={},
y grid style={darkgray176},
ymajorgrids,
ymin=-23.7399952, ymax=904.3530392,
ytick style={color=black},
ytick={0,200,...,800},
yticklabels={}
]
\addplot [line width=2.4pt, darkcyan0101189]
table {%
294.357452392578 693.704162597656
282.264343261719 683.231018066406
270.498260498047 672.391784667969
259.090118408203 661.176513671875
248.078369140625 649.571960449219
237.474380493164 637.593505859375
227.267883300781 625.274597167969
217.46305847168 612.633605957031
208.084228515625 599.673522949219
199.15087890625 586.402465820312
190.67301940918 572.835876464844
182.654632568359 558.99267578125
175.10578918457 544.888061523438
168.068496704102 530.521484375
161.541015625 515.916076660156
153.139831542969 495.586212158203
121.529052734375 417.770111083984
104.299789428711 375.126007080078
99.5452117919922 361.962097167969
95.9270935058594 350.524322509766
92.8480987548828 338.929992675781
90.7429428100586 329.156646728516
89.0727996826172 319.299407958984
87.83984375 309.377746582031
87.0448608398438 299.411315917969
86.6921005249023 289.419311523438
86.7858200073242 279.421569824219
87.3244552612305 269.437866210938
88.5451583862305 257.502685546875
90.368766784668 245.644500732422
92.7745132446289 233.890518188477
95.7506866455078 222.267852783203
99.2866287231445 210.803070068359
103.382522583008 199.526336669922
107.237976074219 190.301406860352
111.509033203125 181.261474609375
116.198417663574 172.431289672852
121.289894104004 163.826644897461
127.888816833496 153.806900024414
134.976119995117 144.126281738281
142.525894165039 134.80192565918
150.533462524414 125.867713928223
157.554733276367 118.74974822998
164.887939453125 111.953567504883
172.519424438477 105.494110107422
182.041000366211 98.1948394775391
191.924713134766 91.3938751220703
202.143035888672 85.1068344116211
212.671310424805 79.3539123535156
223.482177734375 74.1514358520508
234.546401977539 69.512092590332
245.833511352539 65.4448699951172
257.3134765625 61.9589424133301
268.957366943359 59.0674781799316
280.735382080078 56.783275604248
292.616027832031 55.1127738952637
304.566955566406 54.0564041137695
316.556335449219 53.615047454834
328.552642822266 53.7859916687012
340.524291992188 54.5729522705078
352.438812255859 55.9813423156738
364.266845703125 57.9908142089844
375.984893798828 60.5660743713379
387.570617675781 63.6829223632812
399.000549316406 67.3296585083008
410.250946044922 71.4972076416016
421.302215576172 76.16748046875
432.139587402344 81.3150482177734
442.752960205078 86.9099731445312
453.138244628906 92.9179534912109
464.963897705078 100.406723022461
476.477416992188 108.367263793945
489.286743164062 117.950119018555
504.939392089844 130.395401000977
545.148742675781 163.359878540039
672.966125488281 269.253326416016
815.827880859375 388.330505371094
829.302612304688 400.261169433594
840.915649414062 411.263153076172
850.679077148438 421.293273925781
858.694396972656 430.221038818359
866.349182128906 439.459655761719
873.6181640625 449.004791259766
880.487182617188 458.841735839844
886.948791503906 468.950805664062
892.980529785156 479.321685791016
898.539733886719 489.953216552734
903.583984375 500.838592529297
908.083679199219 511.960113525391
912.018493652344 523.293823242188
915.3720703125 534.813110351562
918.137329101562 546.487670898438
920.31005859375 558.286865234375
921.878295898438 570.181457519531
922.830505371094 582.14111328125
923.163635253906 594.134033203125
922.883422851562 606.128295898438
921.993530273438 618.092834472656
920.486389160156 629.995239257812
918.751159667969 639.841735839844
916.577331542969 649.600769042969
913.966674804688 659.252136230469
910.925842285156 668.776672363281
906.72802734375 680.015808105469
901.964233398438 691.027099609375
896.667419433594 701.792236328125
890.860717773438 712.291137695312
884.548645019531 722.493957519531
878.892578125 730.738525390625
872.873352050781 738.721801757812
866.499267578125 746.424743652344
859.784912109375 753.832946777344
852.748840332031 760.936340332031
843.911254882812 769.050476074219
834.67724609375 776.71044921875
825.070434570312 783.897216796875
815.109191894531 790.584106445312
804.817443847656 796.750244140625
794.22607421875 802.386291503906
783.365478515625 807.484313964844
772.259826660156 812.023620605469
762.834594726562 815.359741210938
753.27099609375 818.275512695312
743.587829589844 820.765563964844
731.840026855469 823.201171875
719.985412597656 825.048095703125
708.053466796875 826.301086425781
698.071838378906 826.873840332031
688.07470703125 827.001281738281
678.082214355469 826.676452636719
666.124694824219 825.706970214844
654.22607421875 824.176879882812
640.422668457031 821.859191894531
620.803527832031 817.990966796875
489.638458251953 790.644348144531
474.11572265625 786.775268554688
458.74365234375 782.345642089844
443.539672851562 777.369140625
428.506774902344 771.897216796875
413.651306152344 765.96044921875
398.994171142578 759.5498046875
384.561492919922 752.648803710938
370.378021240234 745.249145507812
356.457244873047 737.366394042969
342.815673828125 729.009765625
329.471099853516 720.186584472656
316.425384521484 710.927124023438
303.690216064453 701.244995117188
294.357452392578 693.704162597656
};
\draw[draw=black,very thick] (axis cs:269.576116702882,209.035930434789) circle (100);
\draw[draw=black,very thick] (axis cs:744.812998936186,668.008000287832) circle (100);
\addplot [line width=2.4pt, black]
table {%
326.107788085938 655.879577636719
262.607147216797 731.528747558594
};
\addplot [line width=2.4pt, black]
table {%
294.357452392578 693.704162597656
282.264343261719 683.231018066406
270.498260498047 672.391784667969
259.090118408203 661.176513671875
248.078369140625 649.571960449219
237.474380493164 637.593505859375
227.267883300781 625.274597167969
226.01985168457 623.711975097656
};
\path [draw=black, fill=black]
(axis cs:193.73149836782,582.951205953849)
--(axis cs:201.26217217231,640.772386779308)
--(axis cs:222.426509917231,624.007220277931)
--(axis cs:223.668253917231,625.574797277931)
--(axis cs:228.371440082769,621.849204722069)
--(axis cs:227.129696082769,620.281627722069)
--(axis cs:248.29403382769,603.516461220692)
--cycle;
\draw (axis cs:269.576116702882,209.035930434789) node[
  text=black,
  rotate=0.0
]{T1};
\draw (axis cs:744.812998936186,668.008000287832) node[
  text=black,
  rotate=0.0
]{T2};
\end{axis}

\end{tikzpicture}
	\end{minipage}
	\begin{minipage}[t]{0.62\columnwidth}
		\setlength{\figH}{2.8cm}
		\input{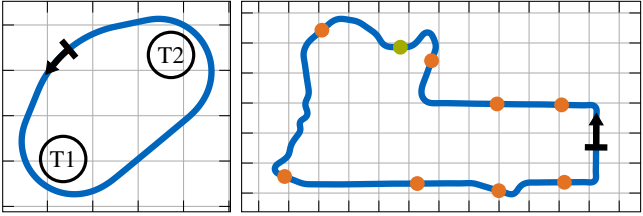}
	\end{minipage}
	\caption{Race tracks of \acs{LVMS} (left) and \acs{MPCB} (right). The arrows indicate the direction of travel and the respective $s=\SI{0}{\meter}$ position. The orange and green dots on the \acs{MPCB} indicate the positions of the static objects mentioned in Section~\ref{sec:racing_line_replanning}. One grid cell equals \SI{200}{\meter} in width and height.}
	\label{fig:lvms_mpcb}
\end{figure}

All experiments are conducted in simulations on the \acf{LVMS} and the \acf{MPCB}, both depicted in Fig.~\ref{fig:lvms_mpcb}. The \ac{LVMS} is a \SI{2400}{\meter} long oval race track with turns reaching a maximum banking of \SI{20}{\degree} enabling speeds of \SI{90}{\meter\per\second}. The \ac{MPCB} is \SI{6250}{\meter} long and exhibits a more complex course. It has a maximum elevation difference of \SI{175}{\meter}, a maximum slope of \SI{13}{\degree}, and a maximum banking of \SI{8}{\degree}. The choice of the respective reference line has a decisive influence on the overall behavior. For generated trajectories with a different orientation than the reference line, the velocity profile $v(t)$ is distorted compared to the specified $\dot{s}(t)$ profile, increasing the probability of non-feasibility. Since, in most cases, the racing line is followed or overtaking maneuvers of a similar orientation are performed, we choose the racing line to be driven as the reference line.

\subsection{Computational Details}
\label{sec:implementation}
\begin{table}[t]
	\caption{Experiment parameters}
	\begin{center}
		\begin{tabular}{c c}
			\toprule 
			Parameter & Value \\
			\midrule
			$H_{\mathrm{rl}}$ & \SI{500}{\meter} \\
			$T$ & \SI{3}{\second} \\
			$N_{\mathrm{\dot{s}}}$ & \num{40} \\
			$K_{\mathrm{\dot{s}}}$ & \num{1.2} \\
			$N_{\mathrm{n}}$ & \num{15} \\
			$s_{\mathrm{0},\mathrm{thr}}$ & \num{0.3} \\
			$d_{\mathrm{w}}$ & \SI{1.93}{\meter} \\
			$d_{\mathrm{s}}$ & \SI{0.2}{\meter} \\
			$\hat{\kappa}_{\mathrm{max}}$ & \SI{0.1}{\per\meter} \\
			$d_{\mathrm{s},\mathrm{rl}}$ & \SI{0.5}{\meter} \\
			\bottomrule
		\end{tabular}
		\qquad
		\begin{tabular}{c c}
			\toprule 
			Parameter & Value \\
			\midrule
			$\tilde{a}_{\mathrm{mgn}}$ & \num{0.1} \\
			$w_{\mathrm{n}}$ & \num{0.1} \\
			$w_{\mathrm{v}}$ & \num{100} \\
			$w_{\mathrm{pr}}$ & \num{5000} \\
			$k_{\mathrm{s}}$ & \num{0.015} \\
			$k_{\mathrm{n}}$ & \num{0.5} \\
			$d_{\mathrm{snr}}$ & \SI{200}{\meter} \\
			$N$ & \num{30} \\
			$\tilde{a}_{\mathrm{abs},\mathrm{mgn}}$ & \SI{0.8}{\meter\per\second\squared} \\
			$\Delta t_{\mathrm{sim}}$ & \SI{0.1}{\second} \\
			\bottomrule
		\end{tabular}
		\label{tab:parameters}
	\end{center}
\end{table}
The reference line $\mathcal{C}$ and the track widths $n_{\mathrm{l}}(s)$ and $n_{\mathrm{r}}(s)$ are available as presampled curve points. For an evaluation between the discrete points, we rely on linear interpolation. The generated trajectories are represented by $N$ equidistantly distributed discrete points in time, and the feasibility checks are performed only for these points. The offline-calculated extremal values and the shape factor for the representation of the gg-diagrams are stored as lookup tables for discrete $v$ and $\tilde{g}$ pairs and are interpolated online. Further, we approximate the cost function in \eqref{eq:cost:function} using the rectangle rule, with the $N$ discrete points along the trajectory being used as integration points.

For racing line generation, we use the procedure proposed in \cite{Rowold2023}. It allows both the offline generation of a closed racing line for the entire race track and the online generation for the desired spatial horizon $H_{\mathrm{rl}}$. Furthermore, it uses the same feasibility checks as in Section~\ref{sec:feasibility_checks}, allowing the racing line to be traversed. The generated racing line is only available as discrete points and is evaluated at the needed $N$ time points. However, due to the necessary interpolation, the resulting points may violate the feasibility checks, particularly affecting checks \eqref{eq:diamond1} to \eqref{eq:diamond3}. If the absolute acceleration limits are low, the relative margin $\tilde{a}_{\mathrm{mgn}}$ introduced in Section~\ref{sec:feasibility_checks} may not cover the violation, as the resulting surplus of permitted limits is also low. For this reason, we use an additional absolute margin $\tilde{a}_{\mathrm{abs},\mathrm{mgn}}$, which guarantees a minimum margin regardless of the absolute limit values.

For simulation, an exact prediction of the opposing vehicles is utilized. The racing line generation, prediction, and local planning are executed sequentially, neglecting the effects resulting from different execution times. After a local planning cycle, the ego and the opposing vehicles are moved along the planned trajectory and prediction by a fixed time $\Delta t_{\mathrm{sim}}$, ensuring a deterministic simulation.\footnote{An implementation of the concept according to the stated details can be found in \url{https://github.com/TUMRT/sampling_based_3D_local_planning}.} All parameters used for the experiments are listed in Table~\ref{tab:parameters}. 

\subsection{Relative Trajectory Generation}
\label{sec:relative_trajectory_generation}
\begin{table}[t]
	\caption{Lap time comparison using different motion primitives}
	\begin{center}
		\begin{tabular}{c c c c} 
\toprule 
& & \multicolumn{2}{c}{\textbf{Lap time in \si{\second}}} \\ 
\cmidrule{3-4} 
\textbf{\#} & \textbf{Mode} & \ac{LVMS} & \ac{MPCB} \\ 
\midrule 
1 & Offline racing line & \num{27.12} & \num{125.96} \\ 
2 & Jerk-optimal trajectory generation & \num{27.12} & \num{137.07} \\ 
3 & Relative trajectory generation & \num{27.12} & \num{125.96} \\ 
\bottomrule 
\end{tabular}
		\label{tab:laptimes_motionprimitives}
	\end{center}
\end{table}

In the first experiment, we compare the performance for a flying solo lap using the jerk-optimal and the proposed relative trajectory generation. Therefore, the objective is to follow the offline-calculated racing line without considering other vehicles. The resulting lap times compared with the lap time of the racing line are shown in Table~\ref{tab:laptimes_motionprimitives}. While the lap time of the racing line is reached on the \ac{LVMS} with both trajectory generation options, the lap time is significantly higher when using the jerk-optimal trajectory generation on the \ac{MPCB}. This is due to the more complex track geometry of the \ac{MPCB} and the associated racing line that the jerk-optimal trajectories cannot reproduce. Note that an entire lap without feasibility violations using the jerk-optimal edges on the \ac{MPCB} is not possible, e.g., due to the poor racing line tracking and the correspondingly high speeds at turn entrances. Therefore, the soft check concept introduced in Section~\ref{sec:feasibility_checks} becomes effective.

\subsection{Planning in \ac{2D} and \ac{3D}}
\label{sec:2d_3d}
Including \ac{3D} effects enables the planning of driving maneuvers that would not be feasible in \ac{2D}. For comparison, we examine an exemplary scenario on the \ac{LVMS}, in which an opposing vehicle drives on the racing line path at a constant speed of \SI{70}{\meter\per\second} and is to be overtaken. This scenario is executed once considering the \ac{3D} effects and once assuming a flat \ac{2D} race track. 

\setlength{\figH}{0.53\columnwidth}
\setlength{\figW}{1.0\columnwidth}
\begin{figure}[t]
	\centering
	\small
	\input{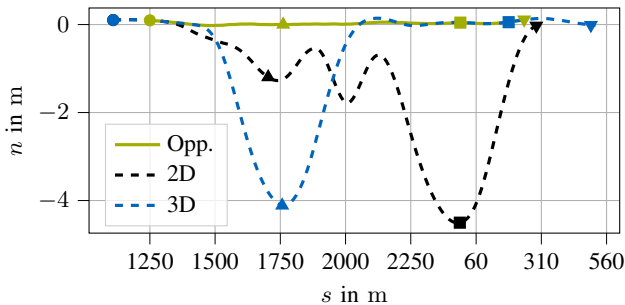}
	\caption{Comparison of an overtaking maneuver on the \ac{LVMS} in curvilinear coordinates using \ac{2D} and \ac{3D} friction checks. The opposing vehicle (Opp.) follows the racing line path ($n\approx\SI{0}{\meter}$) at a constant speed of \SI{70}{\meter\per\second}. Markers of the same shape represent the same point in time.}
	\label{fig:2d_vs_3d}
\end{figure}

Fig.~\ref{fig:2d_vs_3d} shows the overtaking maneuvers performed for both cases in curvilinear coordinates. In the \ac{3D} case, the overtaking maneuver is performed at the first opportunity in turn T2. The banking of the turn causes an increase in $\tilde{g}$ and thus also in the limits, which provides sufficient dynamic potential for the overtake. For the \ac{2D} case, however, a flat track ($\mu=\varphi=\Omega_{\mathrm{x}}=\Omega_{\mathrm{y}}=0$) is assumed in the friction checks, which leads to $\tilde{a}_{\mathrm{x}} = \hat{a}_{\mathrm{x}}$, $\tilde{a}_{\mathrm{y}} = \hat{a}_{\mathrm{y}}$, and $\tilde{g} = g$ according to \eqref{eq:hat_tilde}. This causes the limits to be underestimated in the turn, causing the opposing vehicle to be followed and the overtaking maneuver to be carried out on the main stretch after the turn. The lateral oscillations right before the maneuver result from planning differences within subsequent planning cycles due to the limited planning horizon. 

In addition to neglected dynamic potential, shown exemplarily in this section, the acceleration limits can also be overestimated in the \ac{2D} case. This leads to a violation of the actual limits and occurs, e.g., when traversing a hill while $\Omega_{\mathrm{y}}>0$, resulting in $\tilde{g}<g$ and thus lower limits.

\subsection{Online Racing Line Generation}
\label{sec:racing_line_replanning}
With online racing line generation, the time-optimal trajectory for the finite spatial horizon $H_{\mathrm{rl}}$ is replanned from the current vehicle state in each planning cycle. We compare this approach with the use of an offline-calculated closed racing line that does not consider the current vehicle state.

\begin{table}[t]
	\caption{Lap time comparison using offline and online racing line}
	\begin{center}
		\begin{tabular}{c c c c} 
\toprule 
& & \multicolumn{2}{c}{\textbf{Lap time in \si{\second}}} \\ 
\cmidrule{3-4} 
\textbf{\#} & \textbf{Mode} & \ac{LVMS} & \ac{MPCB} \\ 
\midrule 
1 & Offline racing line (\#1 in Table \ref{tab:laptimes_motionprimitives}) & \num{27.12} & \num{125.96} \\ 
2 & Solo with offline racing line (\#3 in Table \ref{tab:laptimes_motionprimitives}) & \num{27.12} & \num{125.96} \\ 
3 & Solo with online racing line & \num{27.17} & \num{126.13} \\ 
4 & Multi with offline racing line & \num{27.74} & \num{128.38} \\ 
5 & Multi with online racing line & \num{27.28} & \num{126.79} \\ 
\bottomrule 
\end{tabular}
		\label{tab:laptimes_online}
	\end{center}
\end{table}

\setlength{\figW}{1.0\columnwidth}
\begin{figure}[t]
	\centering
	\small
	\input{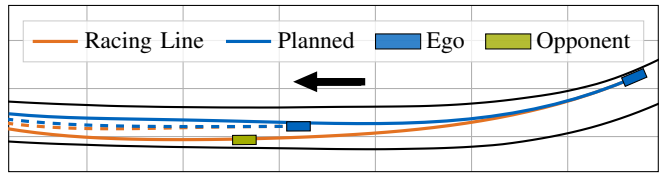}
	\caption{Overtaking maneuver of a static obstacle (green dot in Fig.~\ref{fig:lvms_mpcb}) with online racing line generation on the \ac{MPCB}. The solid and dashed curves correspond to two different planning cycles over time.}
	\label{fig:online_overtake}
\end{figure}

First, we examine a flying solo lap that requires no deviation from the racing line. Entries \#2 and \#3 in Table~\ref{tab:laptimes_online} show the resulting lap times for the \ac{LVMS} and \ac{MPCB}. As already shown in Table~\ref{tab:laptimes_motionprimitives}, when using the offline-generated racing line, the lap times of the racing lines can be retained for both tracks due to the relative trajectory generation. However, the lap times increase with the use of the online racing line generation, as the global time-optimal solution is not sufficiently approached due to the limited spatial horizon.

Subsequently, we consider flying laps while opposing vehicles occupy the racing line, forcing a deviation from it. On the \ac{LVMS}, dynamic objects are placed every \SI{200}{\meter} along the racing line, following it with \SI{70}{\percent} of the actual racing line speed. On the \ac{MPCB}, we consider nine static objects placed along the offline-generated racing line, as depicted in Fig.~\ref{fig:lvms_mpcb}. We choose static objects for the \ac{MPCB}, as the difficulty of the overtaking maneuver varies depending on the position on the race track, and the use of fixed opponent positions ensures reliable comparability of lap times. Entries \#4 and \#5 in Table~\ref{tab:laptimes_online} again list the lap times using the offline and online-generated racing lines. In contrast to solo driving, leveraging online racing line generation yields reduced lap times on both tracks. The higher lap times associated with the offline-generated racing line primarily result from necessary deviations from the racing line speed to facilitate overtaking maneuvers. This discrepancy arises since the trajectories generated to quickly guide the vehicle back to the global racing line speed are not feasible when approaching the vehicle's limits. In contrast, when using online racing line generation, a racing line is generated that brings the vehicle to the maximum speed in the shortest duration, considering the current state. Since the relative trajectory generation is based on this racing line, a return to maximum speed is faster than using offline racing line generation.

Fig.~\ref{fig:online_overtake} illustrates an exemplary overtaking maneuver on the \ac{MPCB}, showcasing the interplay between local trajectory planning and online racing line generation. In the first planning cycle considered, following the racing line would lead to a collision with the static object, causing an evasive maneuver to be planned. In the subsequent planning cycle, which is examined here, the time-optimal racing line corresponds to a shorter path. However, using the racing line generated offline would guide the vehicle away from the actual time-optimal line after overtaking, increasing the required time for this turn.
	\section{Conclusion and Outlook}
We have presented a sampling-based local trajectory planning approach for \ac{3D} race tracks. In contrast to existing approaches based on jerk-optimal edges, the proposed relative trajectory generation allows the lap time of the racing line to be maintained even for complex circuits. The \ac{3D} effects are incorporated by gg-diagrams that depend on the vertical acceleration and ensure that the dynamic potential is not exceeded or underestimated. Finally, we have demonstrated that online racing line generation, in combination with the local planner, slightly increases lap times in the solo scenario but can effectively reduce them for the multi-vehicle scenarios evaluated, especially on complex race tracks.

In the future, we will evaluate the proposed planning approach on a full-scale vehicle as part of the software stack of the TUM Autonomous Motorsport team. For this, the combination of local planner and control algorithm must be investigated with regard to the vehicle stability. Additionally, we will focus on greater maneuver diversity. This includes generating trajectories that differ from the jerk-optimal or relative trajectories to perform more complex maneuvers. It could also reduce the required margin of the limits between the sampling-based planner and the racing line generation. Another aspect is the addition of a second sampling stage, which would make it possible to plan outward and inward maneuvers within a single planning cycle. Finally, the effectiveness of online racing line generation should be investigated further. This could involve more complex scenarios and the use of alternative local planning approaches.
	
	\bibliographystyle{IEEEtran}
	\bibliography{IEEEabrv, references}

\end{document}